\pdfoutput=1
\documentclass[11pt]{article}

\usepackage[preprint]{acl}

\usepackage{times}
\usepackage{latexsym}
\usepackage{framed}

\usepackage[T1]{fontenc}
\usepackage[utf8]{inputenc}

\usepackage{microtype}

\usepackage{inconsolata}

\usepackage{graphicx}
\usepackage{float}
\usepackage{stfloats}

\title{Retrieval-Augmented Generation for Domain-Specific Question Answering: A Case Study on Pittsburgh and CMU}

\author{
  *Haojia Sun \\
  Carnegie Mellon University \\
  \texttt{haojias@andrew.cmu.edu} \\\And
  *Yaqi Wang \\
  Carnegie Mellon University \\
  \texttt{yaqiwang@andrew.cmu.edu} \\\And
  *Shuting Zhang \\
  Carnegie Mellon University \\
  \texttt{shuting2@andrew.cmu.edu} \\}

\begin{document}
\maketitle
\begin{abstract}
% We proposed the design and implementation of a Retrieval-Augmented Generation (RAG) system tailored to provide large language models with relevant documents for answering domain-specific questions related to Pittsburgh and Carnegie Mellon University (CMU). Our approach encompasses a comprehensive data extraction process, utilizing a greedy scraping strategy to collect over 1,800 subpages from diverse sources such as Wikipedia, city government websites, and event calendars. We employed a hybrid data annotation methodology combining manual annotation and Mistral-generated question-answer pairs, achieving a high inter-annotator agreement (IAA) score of 0.7625. The RAG framework integrates both BM25 and FAISS retrievers, enhanced with a reranker to optimize document retrieval accuracy. Experimental evaluations demonstrated that the RAG system significantly outperformed a baseline non-RAG model, particularly excelling in time-sensitive and complex queries with an F1 score improvement from 5.45\% to 42.21\%, a recall score as 56.18\%. Our findings highlight the effectiveness of retrieval augmentation in enhancing the precision and relevance of generated answers, while also identifying areas for improvement in document retrieval and model training. This work underscores the potential of RAG systems in specialized domains, offering a robust foundation for future advancements in knowledge-enhanced language models.
We designed a Retrieval-Augmented Generation (RAG) system to provide large language models with relevant documents for answering domain-specific questions about Pittsburgh and Carnegie Mellon University (CMU). We extracted over 1,800 subpages using a greedy scraping strategy and employed a hybrid annotation process, combining manual and Mistral-generated question-answer pairs, achieving an inter-annotator agreement (IAA) score of 0.7625. Our RAG framework integrates BM25 and FAISS retrievers, enhanced with a reranker for improved document retrieval accuracy. Experimental results show that the RAG system significantly outperforms a non-RAG baseline, particularly in time-sensitive and complex queries, with an F1 score improvement from 5.45\% to 42.21\% and recall of 56.18\%. This study demonstrates the potential of RAG systems in enhancing answer precision and relevance, while identifying areas for further optimization in document retrieval and model training.
\end{abstract}

\section{Introduction}
In this report, we present our approach to designing and implementing a Retrieval-Augmented Generation(system) that provides large language models with retrieved relevant document to answer the questions for specific domains. Our focus is on the questions related to Pittsburgh and Carnegie Mellon University(CMU), covering a broad range of topics including history, events, culture and sports.

In the second section, we illustrated how we conducted data extraction process, sourcing information from a wide range of publicly available websites such as Wikipedia, city government websites, and event calendars and greedily extracting their subpages. In the third section, we discussed how we cominbined Mistral and human annotation for data annotaion. We used  inter-annotator agreement(IAA) to evaluate the quality of data annotation. In the fourth section, we discussed our design of RAG system and how we turn the documents into vector database. In the next section, we illustrated how we conducted the experiments. Finally, we showed and analyzed the results of the RAG system on our test data.

\section{Data Extraction}
To compile our knowledge resource for the Retrieval-Augmented Generation (RAG) system, we started by identifying relevant domains specifically related to Pittsburgh and Carnegie Mellon University (CMU). This included general topics such as history, events, culture, and government information, ensuring that our dataset would cover a broad range of queries. The sources we included ranged from well-known public websites like Wikipedia and the City of Pittsburgh official website, to more niche resources like event calendars, and food festival pages. In total, we collected approximately 61 webpages and relevant documents, classifying them into distinct categories such as government, city information, sports, food, culture, museums, music, events, history, and school. \par

To collect subpages, we employed a greedy scraping strategy, defining specific keywords such as "pirates","picklesburgh" and "cmu.edu/about/" that must be included in the corresponding subpages for each webpage to target relevant subpages automatically. We also defined the unwanted words for each webpage such as "news","stats","instagram", and "youtube" to ensure the unwanted subpages were not included. The web crawler was designed to run a Breadth-First Search (BFS) to explore subpages recursively, ensuring that all essential subpages linked to the main websites were captured while avoiding irrelevant or redundant pages. We altogether collected around \textbf{1820 subpages, 7 PDFs, and 16 tables}. \par

\begin{figure*}[t]
  \includegraphics[width=0.9\textwidth]{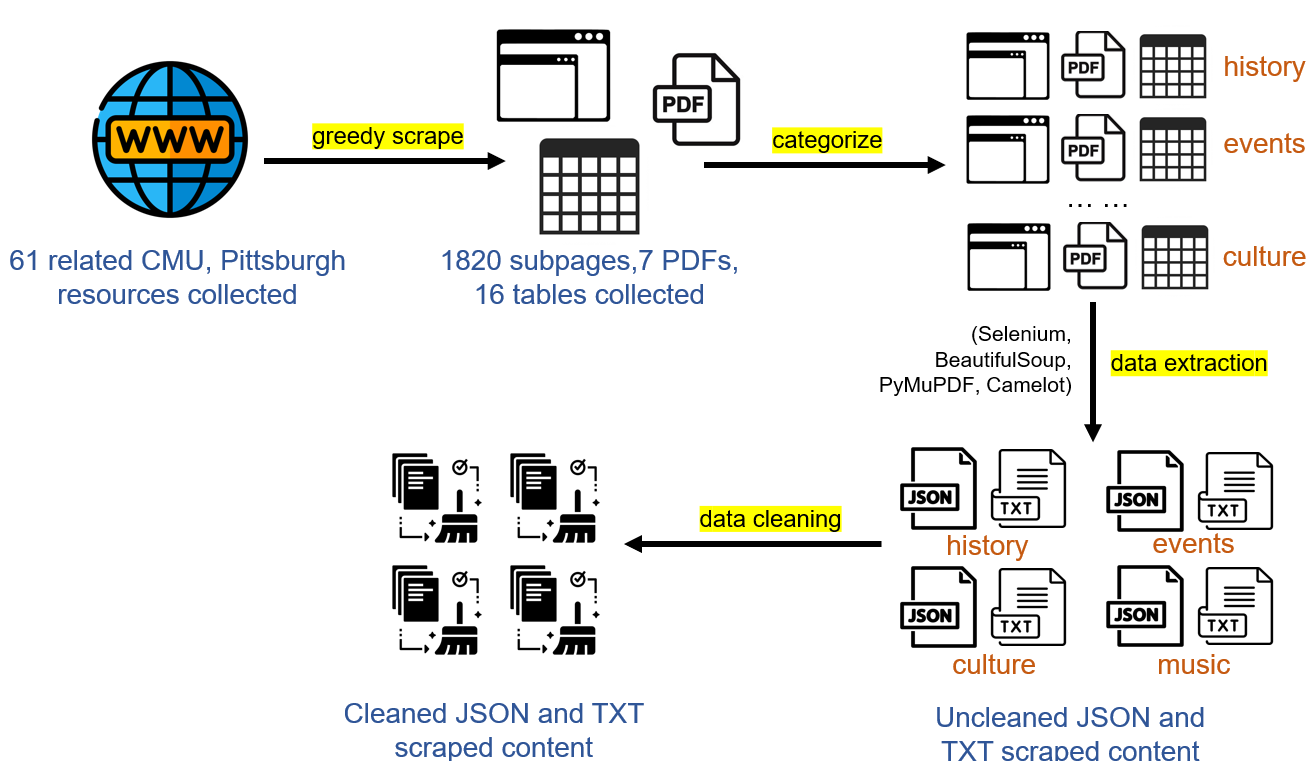}
  \caption{Design of Data Extraction.}
  \label{fig:data_creation}
\end{figure*}

Figure \ref{fig:data_creation} shows the whole pipeline for data extraction. For data extraction, we used several tools to handle different file formats and content types. For webpages, we relied on \textbf{Selenium} and \textbf{BeautifulSoup} to automate browsing, and parse the HTML. For handling PDFs and tables, we employed \textbf{PyMuPDF} and \textbf{Camelot}. Selenium’s headless browser was configured with appropriate headers, including User-Agent and cookies, to mimic a regular browser and ensure proper scraping. We specifically extracted content from tags like <h2>, <h3>, <h4>, <p>, <div>, <span>, and <article>, ensuring that all significant textual data was captured. To prevent scraping the same pages multiple times, we employed a set-based tracking mechanism where every visited URL was stored, ensuring we didn’t duplicate our efforts. Additionally, by filtering out subpages with less than 200 characters or those with generic titles like "Page not found," we ensured that our dataset contained high-quality content that was relevant to the queries our system would need to answer. \par

\section{Data Annotation}
\begin{figure*}[h]
  \centering
  \includegraphics[width=0.9\textwidth]{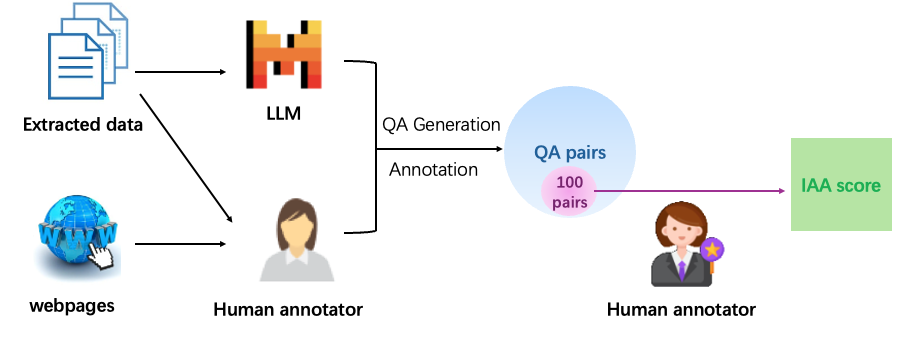}
  \caption{Pipeline of Data annotation}
  \label{fig:data_annotation}
\end{figure*}

Our data annotation process aimed to generate a diverse and representative set of question-answer (QA) pairs that would serve as test data for evaluating our RAG system. We used a combination of \textbf{manual annotation} and \textbf{automatic generation} via an open-source model (Mistral \cite{jiang2023mistral7b}) to ensure the robustness and completeness of our annotated dataset. Figure \ref{fig:data_annotation} shows the whole pipeline for data annotation. \par

To begin, we generated a total of \textbf{1,467} QA pairs, of which 165 pairs were manually created by randomly selecting content from the resources we scraped. The remaining 1,302 QA pairs were automatically generated using Mistral in a combination of few-shot learning and fine-tuning. Each QA pair was also labeled for \textbf{time sensitivity}, with a binary label (0 or 1) indicating whether the answer was influenced by temporal factors. Each QA pair was sourced from one of two types of input:
\begin{itemize}
    \item \textbf{Scraped content}: This included documents, web pages, and text from city websites, event pages, and other sources related to Pittsburgh and CMU.
    \item \textbf{Randomly selected content}: We manually extracted random segments of text from the collected resources to ensure diverse coverage, especially for validating the completeness of the data.
\end{itemize}

We decided what and how much data to annotate with two main goals: diversity and completeness.
For manual generation, we manually selected 165 segments from different categories of our scraped resources (e.g., events, history, culture, government information) to ensure broad topic coverage for Pittsburgh and CMU. Each segment was used to write corresponding questions and answers, validating the completeness of our dataset. \par

For model generation, we used Mistral with \textbf{few-shot learning}. Few-shot learning allowed the model to generate more examples based on manually created QA pairs. We provided Mistral with a few examples of manually created QA pairs, which it used to generate additional pairs from our scraped content. This allowed the model to generalize from the small number of examples and apply that knowledge to new text.

% \begin{itemize}
%     \item \textbf{Few-shot learning}: We provided Mistral with a few examples of manually created QA pairs, which it used to generate additional pairs from our scraped content. This allowed the model to generalize from the small number of examples and apply that knowledge to new text.
%     \item \textbf{Fine-tuning}: We fine-tuned Mistral on the manually annotated data (165 QA pairs) to improve its performance in generating high-quality, domain-specific QA pairs. However, given the relatively small size of this dataset, it was not sufficient to train the model to a high level of accuracy for domain-specific queries related to Pittsburgh and CMU. To address this, we employed a data augmentation strategy to increase the diversity and volume of the training data. Specifically, we employed Back-Translation and Paraphrasing methods to produce varied question and answer formulations while retaining the original meaning of the content. This allowed us to increase the size of the training dataset to approximately \textbf{1200} pairs. We used cross-entropy loss as the primary loss function for the fine-tuning process. We employed a small batch size of \textbf{16} to ensure stable gradient updates and used a learning rate of \textbf{2e-5}, and we adjusted the learning rate using a linear decay scheduler to avoid overfitting during training. We ran the fine-tuning process for 3 epochs to allow the model to converge. This process helped to refine the model’s ability to handle specific types of queries related to Pittsburgh and CMU.
% \end{itemize}

We aimed for a balanced approach between manual and model-generated pairs to ensure diversity and consistent quality across topics. The manually generated pairs established a strong base for testing, while the model-generated pairs expanded coverage across our scraped data.

\subsection{Data Quality Estimation}
To evaluate the quality of our annotated dataset, we calculated the inter-annotator agreement(IAA) using the F1 score. This metric is used to measure partial overlaps in the answers annotated by model or by different people and F1 score balances the precision and recall to quantify the agreement between different ways of annotations.

To obtain the IAA score, we randomly selected 100 questions from the annotations given by either model and human annotators. The selected questions cover all the four areas including history, events, food and culture, and sports. Human annotator manually annotated the questions for comparison between answers. We calculated F1 score for each answer pairs and averaged the F1 score as the IAA score. 

The IAA score calculated is \textbf{0.7625}, which represents a generally high quality of data annotation. Some disagreements occur in the data format such as "Iron City" and "The Iron City" and some occur in the different numbers such as "percentage allocation of the Director is funded by the Department of Finance in 2024". Another kind of disagreement lies in the errors of answers such as answering the time to "where" questions.

\section{RAG Framework Design}
This section we will discuss the backbone LLM model selection, retriever model selection and how we created a vector database for better retrieval, showing the details of using LangChain\cite{langchain2023framework} to build a RAG pipeline.

\subsection{Overview Design}
%digram
\begin{figure*}[t]
  \includegraphics[width=\textwidth]{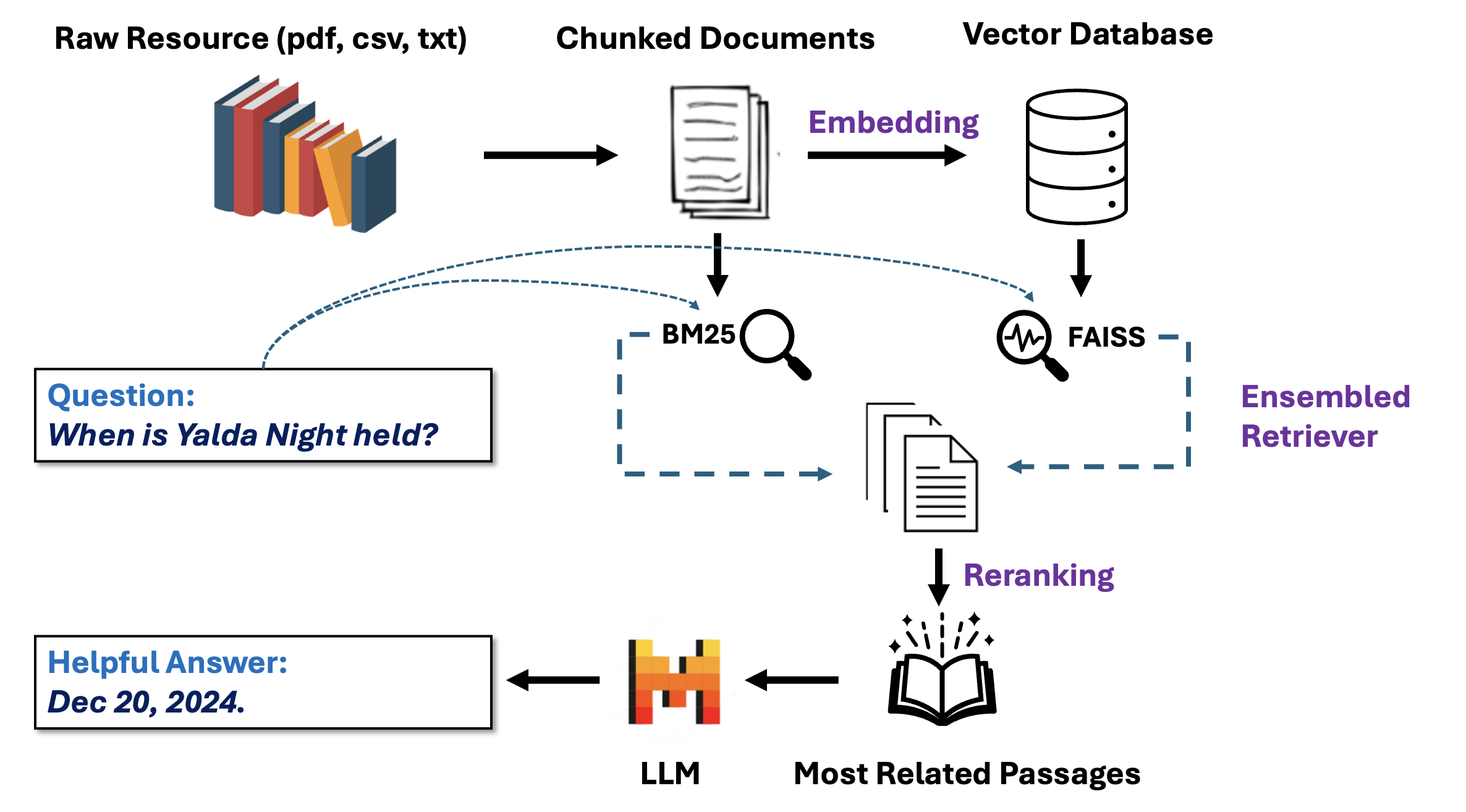}
  \caption{Design of RAG workflow}
  \label{fig:workflow}
\end{figure*}

As shown in Figure~\ref{fig:workflow}, the first step involves building chunked documents. After extracting data from various websites, we obtain different types of files, including PDFs, CSVs, and TXT formats. We convert all of these into simple text and recursively chunk them into smaller pieces, with a chunk size of 1000. To avoid losing important information during this chunking process, we set \texttt{chunk\_overlap}=200, ensuring that the truncated part of the text is preserved.

Secondly, we choose to combine both BM25\cite{robertson2009bm25} and FAISS\cite{johnson2017faiss} retrievers. FAISS is designed for quickly searching through large volumes of documents, though it may sacrifice some accuracy. BM25 helps compensate for this by improving precision. During our experiments, we observed that including too many unrelated documents negatively impacted the RAG generator’s performance. To address this, we added a reranker module. We selected a small model, sentence-transformers’ all-MiniLM-L6-v2\cite{jiang2023mistral7b}, which is only 90.9 MB, ensuring the overall process remains efficient.

Finally, for the backbone LLM, we opted for a 7B Mistral model\cite{jiang2023mistral7b}. By incorporating related documents into the prompt and employing 2-shot learning, we successfully generate the desired result.

\subsection{Vector Database Creation}
Document database begins by first loading the documents from various file types, such as CSV, JSON, and TXT. Depending on the file type, a specific loader is used. For JSON files, the content is converted into markdown format to ensure better structuring, while the other file types are directly read and loaded into memory as Document objects.

Once the documents are loaded, if their content size exceeds a specified threshold, the documents are split into smaller chunks using the \texttt{RecursiveCharacterTextSplitter}. This ensures that the document contents are manageable for processing without losing important information. Overlap between chunks is controlled by the \texttt{chunk\_overlap} parameter to prevent information loss during the chunking process.

Next, embeddings for the chunked documents are generated using a Embedding model (sentence-transformers/multi-qa-mpnet-base-dot-v1 \cite{jiang2023mistral7b}), which converts the textual information into numerical vectors. These vectors are then used to index the documents in FAISS, a vector similarity search engine. The FAISS retriever is constructed from the embedded documents and saved locally for future retrieval tasks.

\section{Experiments}
In this section, we will first introduce the baselines we chosen. We use a small QA dataset for hyper-parameter tuning for RAG model, and then we compare the best RAG model with the naive LLM model.

\subsection{Experiments Setup}
During the hyperparameter tuning process for the RAG model, several key parameters were carefully adjusted to optimize model performance. These include the maximum number of new tokens generated, the number of top retrieved documents for each retriever(bm25 and FAISS), and the number of documents remained after reranking. Additionally, we evaluated the impact of using a re-ranker, few shot learning as well as the ensembled retriever, comparing their performance against setups without these features. We set the backbone model's \texttt{top\_p = 1}, which means every token is chosen as the token with highest probability, enabling the reproduction ability.

\subsection{Metrics}
To evaluate the performance of the RAG model, we utilized several key metrics, including Exact Match (EM), Precision, Recall, and F1 Score. The Exact Match measures the percentage of predictions that match the reference answer exactly, while Precision is the ratio of correctly predicted tokens to the total number of predicted tokens. Recall represents the ratio of correctly predicted tokens to the total number of tokens in the reference answer, and F1 Score is the harmonic mean of Precision and Recall.

Following the evaluation scripts of SQuAD\cite{rajpurkar2016squad}, we normalized the reference and generated answers by lowercasing the text and removing punctuation, articles, and extra whitespace. This ensured consistency when comparing predictions. The evaluation process included tokenizing both the reference and generated answers, calculating common tokens, and using this information to compute Precision, Recall, and F1 scores. For cases where either answer was missing, the F1 score was adjusted accordingly, with special handling for no-answer scenarios.

\subsection{Results}
In the experiments, we explored ariations with and without document re-ranker, few-shot learning, and an ensembled retriever (combining BM25 and FAISS). The performance was measured using \textbf{Exact Match (EM), Precision, Recall, and F1 Score}, which are standard metrics for evaluating question-answering systems. As shown in Table \ref{result_table}, we evaluated eight different setups of the RAG framework.

\textbf{Baseline (No RAG).} The baseline model, without retrieval augmentation, resulted in poor performance, with an EM of 0.00\% and an F1 score of 5.45\%. This highlights the model's inability to generate accurate answers without the aid of external documents.

\textbf{RAG without re-ranker or few-shot learning.} Introducing RAG with no document re-ranker and no few-shot learning resulted in an EM of 2.00\% and a modest improvement in F1 score to 19.75\%, indicating the positive effect of retrieval augmentation.

\textbf{Ensembled Retriever.} Using an ensembled retriever (BM25 + FAISS) further improved precision and recall. The best results were achieved with RAG, re-ranker, few-shot learning, and an ensembled retriever, with an EM of 20.25\% and an F1 score of 42.21\%.

\textbf{Few-shot learning impact.} Applying few-shot learning significantly improved the model's performance. When few-shot learning was enabled alongside the ensembled retriever and re-ranker, the F1 score increased to 42.21\%, the highest among all configurations. Without few-shot learning, the F1 score was significantly lower, highlighting the effectiveness of this approach.

\textbf{Document re-ranker.} Document re-ranker slightly reduced the model's ability to recall information but improved precision. This suggests that while smaller, more concise document chunks are beneficial for model precision, they might lead to some loss in recall.

Overall, the best configuration combined RAG with document re-ranker, few-shot learning, and an ensembled retriever, achieving the highest F1 score of \textbf{42.21\%}, an EM of \textbf{20.25\%}, a Precision of \textbf{47.29\%}, and a Recall of \textbf{56.18\%}. These results are statistically significant compared to the baseline and other configurations, demonstrating the effectiveness of the RAG system with few-shot learning and retrieval enhancements.

\begin{table*}[htbp]
\centering
\begin{tabular}{cccccccc}
\hline
\textbf{RAG} & \textbf{Re-ranker} & \textbf{Few-shot} & \textbf{Ensembled Retriever} & \textbf{EM} (\%) & \textbf{Precision (\%)} & \textbf{Recall} (\%) & \textbf{F1} (\%) \\ \hline\hline
No & - & No & - & 0.00 & 3.11 & 33.59 & 5.45 \\ \hline
Yes & No         & No     & No         & 2.00       & 14.25         & 49.25       & 19.75        \\ 
Yes & No         & No     & Yes         & 2.00       & 15.85         & 49.82       & 21.94         \\ 
Yes & No         & Yes     & Yes         & 14.00            & 37.05          & 41.46       & 33.79         \\ 
Yes & Yes        & No     & No         & 0.63            & 12.90       & 47.84     & 18.37       \\ 
Yes & Yes          & No      & Yes          & 0.00             & 14.07          & \textbf{56.18}       & 20.50         \\ 
Yes & Yes         & Yes      & No         & 12.00             & 35.42          & 37.02       & 32.49        \\ 
Yes & Yes         & Yes     & Yes         & \textbf{20.25}            &\textbf{47.29}         & 45.39       & \textbf{42.21}         \\ \hline
\end{tabular}
\caption{Baseline Comparison Results}
\label{result_table}
\end{table*}

\section{Analysis}
% \subsection{Performance across Different Types of Questions}
\subsection{Performance on Time-Sensitive v.s. Non-Time-Sensitive Questions}
In our analysis, we categorized the test set into two main types: time-sensitive and non-time-sensitive questions. Out of the 165 sampled questions, 57 were classified as time-sensitive. The RAG model demonstrated a clear advantage in answering time-sensitive questions, achieving higher accuracy and relevance in its answers compared to the non-RAG model.

This performance difference can be attributed to the retrieval-augmented mechanism of RAG, which allows the model to access up-to-date or contextually relevant information from retrieved documents, whereas the non-RAG model must rely solely on pre-existing knowledge from its training data. In many cases, the non-RAG model's responses were either outdated or vague, especially when specific dates or timely details were required.

For example, consider the following question:

\begin{framed}
    \noindent $\bullet$ \textbf{Question}: When is "Alumni Awards Ceremony" of CMU held at Pittsburgh? \\
    $\bullet$ \textbf{RAG answer}: November 1, 2024, 6:30 pm. \\
    $\bullet$ \textbf{Non-RAG answer}: The Carnegie Mellon University Alumni Awards Ceremony is typically held in the late spring or early summer each year. The exact date varies, so it's best to check the CMU Alumni Association website or contact them directly for the most current information.
\end{framed}

In this example, the RAG model was able to provide a precise and up-to-date answer for the specific event. This is likely because the RAG model retrieved relevant documents or event listings that contained the correct date and time. On the other hand, the non-RAG model provided a more general, vague answer, relying on information that is not specific to the current instance of the event.

\begin{table}[ht]
\centering
\begin{tabular}{ccccc}
\hline
\textbf{Model} & \textbf{TS}  & \textbf{Precision} & \textbf{Recall} & \textbf{F1-Score} \\
\hline
RAG & 0 &  0.3755 & 0.3994 & 0.3559 \\
RAG & 1 &  0.2494 & 0.2335 & 0.2263 \\
\hline
Non-RAG & 0 & 0.0411 & 0.3254 & 0.0704 \\
Non-RAG & 1 &  0.0221 & 0.1668 & 0.0378 \\
\hline
\end{tabular}
\caption{Comparison of RAG and Non-RAG models across different metrics. TS for time-sensitive.}
\label{tab:ts}
\end{table}

As shown in Figure~\ref{tab:ts}, we can see that although the indicators of Non-RAG are generally lower than those of the RAG model, for non-time-sensitive data, its recall is only 18.5\% lower than that of RAG, but for time-sensitive data, its recall is 28.5\% lower than that of RAG. This shows that RAG can help the model to be more up-to-date.

% \subsubsection{Performance on Numerical Questions}

\subsection{Error Source Analysis Based on Retrieved Documents}
In some cases, the RAG model underperformed compared to the non-RAG model, because of the impact of incorrect or misleading document retrieval. These errors can occur due to two primary reasons: wrong document retrieval and interference from pre-trained knowledge. 

A key issue we observed is that when the retrieved documents are not highly relevant to the question, the RAG model can provide answers that are either vague or completely incorrect. This issue is particularly evident when the model retrieves information that contains numeric data or financial terms but does not directly answer the question, which confuses the model. 

For example, consider the following case:
\begin{framed}
    \noindent $\bullet$ \textbf{Question}: What is the FTE (Full-Time Equivalent) for the Director - City Treasurer position in 2024? \\
    $\bullet$ \textbf{RAG answer}: 12 months, \$ budget. \\
    $\bullet$ \textbf{Non-RAG answer}: FTE = 1.04.
\end{framed}

In this case, the RAG model retrieved irrelevant documents that likely contained general budgetary information (e.g., about the length of the financial year or allocation of funds), which led it to incorrectly provide an answer referencing "12 months" and "\$ budget." The model was misled by the retrieved content that did not contain specific FTE information, thus providing an irrelevant answer.

On the other hand, the non-RAG model generated the correct answer, "FTE = 1.04," which suggests that the information required to answer this question was likely learned during pre-training. The non-RAG model's internal knowledge allowed it to correctly answer without external document interference.

\subsection{Impact of Question Length and Complexity on Model Performance}
We categorized the test questions into three groups:
\begin{itemize}
    \item Simple and Short Questions: These questions have fewer than 10 tokens and generally require straightforward, fact-based answers.
    \item Moderately Complex Questions: These questions have between 10-20 tokens and may require reasoning or connecting multiple pieces of information.
    \item Long and Complex Questions: These questions exceed 20 tokens and often involve multi-step reasoning.
\end{itemize}

As the complexity of the questions increased, the performance gap between the RAG and non-RAG models widened. For long and complex questions, the RAG model consistently outperformed the non-RAG model. However, both models demonstrated a higher rate of errors compared to simpler questions. An example of a long and complex question is shown below:

\begin{framed}
    \noindent $\bullet$ \textbf{Question}: Who sent an ultimatum to conductor Emil Paur regarding the hiring of European musicians for the Pittsburgh Symphony Orchestra? \\
    $\bullet$ \textbf{RAG answer}: President Joseph Weber of the National Federation of Musicians. \\
    $\bullet$ \textbf{Non-RAG answer}: 
    Answer: The ultimatum was sent by the Pittsburgh Citizens' Committee for a 100\% American Pittsburgh Symphony Orchestra. This group was led by Harry C. Trexler, a prominent Pittsburgh businessman and civic leader, who believed that the orchestra should only employ American musicians. The ultimatum was delivered to conductor Emil Paur in 1933, and he ultimately gave in to the pressure and dismissed all European musicians from the orchestra. This incident is a significant moment in the history of the Pittsburgh Symphony Orchestra and the broader context of American cultural nationalism during
\end{framed}

In this case, the RAG model was more successful because it retrieved relevant documents that provided the historical context, while the non-RAG model struggled with the multi-step reasoning required to correctly answer both parts of the question.

% \subsection{Effectiveness of the Retrieve-and-Augment Strategy}

\section{Limitations and Future Improvement}
Our study demonstrates the substantial benefits of integrating a Retrieval-Augmented Generation (RAG) system for domain-specific question answering, particularly in contexts requiring up-to-date and detailed information. However, several limitations must be acknowledged. Firstly, the quality of the RAG system is inherently dependent on the relevance and accuracy of the retrieved documents. 

Additionally, the inter-annotator agreement (IAA) score of 0.7625, while indicative of generally high-quality annotations, reveals discrepancies in data formatting and numerical precision. These inconsistencies highlight the challenges in achieving uniformity in annotations, especially when combining manual and automated processes. Future work should explore more refined annotation protocols and possibly incorporate more advanced validation techniques to further enhance data quality.

Finally, the reliance on specific tools and models, such as Mistral and the Mistral 7B model, may limit the system's adaptability to emerging technologies and methodologies. Continuous evaluation and integration of newer models and tools are essential to maintain the system's competitiveness and effectiveness.

\section{Conclusion}
In conclusion, our research presents a successful implementation of a Retrieval-Augmented Generation (RAG) system designed to answer domain-specific questions related to Pittsburgh and Carnegie Mellon University. By combining data extraction, hybrid annotation, and a RAG framework with BM25 and FAISS retrievers, we achieved notable improvements in answer accuracy, especially for time-sensitive and complex queries.

While the system shows strong potential, challenges such as document retrieval accuracy and dataset generalizability remain. Addressing these issues by refining retrieval methods and expanding the dataset will further enhance performance. This work highlights the promise of RAG systems in improving the capabilities of large language models, with future efforts focused on broadening applications and optimizing retrieval processes.

%\bibliography{anthology,main}
\bibliography{main}

% \appendix

% \section{Example Appendix}
% \label{sec:appendix}

% This is an appendix.

\end{document}